\documentclass[10pt, twocolumn]{article}

\usepackage[margin=0.8in]{geometry} 
\usepackage[utf8]{inputenc}
\usepackage[pdftex]{graphicx}
\usepackage{amsmath}
\usepackage{amsfonts}
\usepackage{amssymb}
\usepackage{algorithm}
\usepackage{algpseudocode}
\usepackage{subcaption}
\usepackage{url}
\usepackage{authblk} 

\newcommand{\cut}[1]{}
\newcommand{\mbf}[1]{\mathbf{#1}}

\title{GPU-GLMB: Assessing the Scalability of GPU-Accelerated Multi-Hypothesis Tracking}

\author[1]{Pranav Balakrishnan\thanks{These authors contributed equally to this work.}}
\author[1]{Sidisha Barik\textsuperscript{*}}
\author[2]{Sean M. O'Rourke}
\author[1]{Benjamin M. Marlin}

\affil[1]{Manning College of Information and Computer Sciences, University of Massachusetts Amherst, USA}
\affil[2]{U.S. Army Combat Capabilities Development Command, Army Research Laboratory, Adelphi, MD, USA}

\date{} 

\begin{document}

\maketitle

\renewcommand{\thefootnote}{\fnsymbol{footnote}}
\footnotetext{\copyright~2025 IEEE. Personal use of this material is permitted. Permission from IEEE must be obtained for all other uses, in any current or future media, including reprinting/republishing this material for advertising or promotional purposes, creating new collective works, for resale or redistribution to servers or lists, or reuse of any copyrighted component of this work in other works.}
\renewcommand{\thefootnote}{\arabic{footnote}}

\begin{abstract}
Much recent research on multi-target tracking has focused on multi-hypothesis approaches leveraging random finite sets. Of particular interest are labeled random finite set methods that maintain temporally coherent labels for each object. While these methods enjoy important theoretical properties as closed-form solutions to the multi-target Bayes filter, the maintenance of multiple hypotheses under the standard measurement model is highly computationally expensive, even when hypothesis pruning approximations are applied. In this work, we focus on the Generalized Labeled Multi-Bernoulli (GLMB) filter as an example of this class of methods. We investigate a variant of the filter that allows multiple detections per object from the same sensor, a critical capability when deploying tracking in the context of distributed networks of machine learning-based virtual sensors. We show that this breaks the inter-detection dependencies in the filter updates of the standard GLMB filter, allowing updates with significantly improved parallel scalability and enabling efficient deployment on GPU hardware. We report the results of a preliminary analysis of a GPU-accelerated implementation of our proposed GLMB tracker, with a focus on run time scalability with respect to the number of objects and the maximum number of retained hypotheses.  
\end{abstract}

\section{Introduction}
Multi-object tracking (MOT) is a fundamental capability in numerous domains, including autonomous driving, robotics, and surveillance \cite{blackman1986multiple}. The objective is to estimate the time-varying number of objects and the  individual object states from sequences of sensor observations. This task has many challenges, including sensor noise, detection uncertainty, and the critical problem of data association (correctly linking measurements to tracks over time) \cite{fortmann1980multi, Bar-Shalom2009}.

Within the Bayesian filtering paradigm \cite{stone2013bayesian}, the Random Finite Set (RFS) framework provides a mathematically rigorous foundation for MOT by modeling the multi-object state as a finite set-valued random variable \cite{vo2014labeled}. Among RFS-based methods, the standard Generalized Labeled Multi-Bernoulli (GLMB) filter provides an exact closed-form solution to the multi-target Bayes recursion \cite{WilliamsMaMB2015,Gibbs_Efficient_2017}. A key advantage of the GLMB formulation is its use of labeled RFSs, which naturally incorporates unique and persistent track identities (labels) into the state, thereby yielding estimates of object trajectories directly without the heuristic post-processing needed by earlier multi-hypothesis approaches \cite{ClarkNarykovStreit2023,MahlerBook1,phd-filter}.

However, the GLMB filter's update step requires evaluating a posterior density that is a weighted sum over multiple hypotheses, where each hypothesis corresponds to a specific assignment of measurements to tracks. The number of such hypotheses grows combinatorially with the number of tracks and measurements, rendering an exact implementation of the GLMB filter computationally intractable for real-time applications. To address the computational bottleneck, existing GLMB implementations approximate the posterior by retaining only the most likely hypotheses. These are typically identified using iterative optimization or sampling algorithms \cite{bourgeois1971extension, Multi_sensor_2019}. However, these methods remain inherently sequential, making them a poor match for GPU acceleration \cite{owens2008gpu}.

An additional issue with the standard GLMB filter arises when it is applied in a setting with multiple computer vision-based virtual sensors \cite{pmlr-v216-samplawski23a, sm2025}. This sensing approach combines a network of camera sensors with 2D object detectors \cite{jiang2022review}. The 2D object detections are then projected into world space yielding 3D geospatial detections \cite{sm2025}. However, computer vision-based models have the potential to issue multiple detections for the same object with overlapping or offset bounding boxes \cite{jiang2022review}. These multiple image-space detections then yield multiple geospatial detections for the same object. The potential for multiple detections per object violates the point-target measurement model used in the standard GLMB filter \cite{vo2014labeled}. While frameworks such as Extended Target Tracking (ETT-GLMB) can model multiple detections per object \cite{ETT_2016}, they typically do so by introducing models of object geometry that result in significantly more complex trackers. 

In this paper, we propose a modified GLMB filter that can associate multiple point detections with each object without introducing a model of object geometry. We show that our approach breaks the inter-dependence between detections during the hypothesis generation step of the filter. This allows a batch of detection-to-object associations to be sampled in parallel. We build on this new GLMB formulation to develop a GPU-accelerated tracker, GPU-GLMB. We implement GPU-GLMB using PyTorch, a widely used Python deep learning library that provides acceleration for GPU-bound tensor computations \cite{paszke2019pytorch}.  We begin by discussing related work in Section \ref{related}. We then present the modified GLMB filter in Section \ref{approach} and discuss the GPU-accelerated implementation. We present scaling experiments in Section \ref{experiments}.

\section{Related Work}
\label{related}
As mentioned above, the data association problem, solved across a set of index dimensions (time, filter components, objects, sensors, and/or all of the above), forms the core of MOT. Standard trackers like the Joint Probabilistic Data Association (JPDA) filter \cite{Achkasov1972,FortmannBarShalomScheffeJPDA1983, ORourke2023} and the Multiple Hypothesis Tracker (MHT) \cite{ReidMHT1979, DanchickNewnamMHT2006, ChongMoriReidMHT2018} assume a finite, (mostly) known number of true objects and represent uncertainty in the association between objects and a collection of detections or measurements. In both MHT and JPDA, association hypothesis enumeration presents a combinatorial challenge necessitating the use of $K$-best hypothesis approximation methods for larger-scale problems \cite{DanchickNewnamMHT2006}. 

On the other hand, RFS trackers consider the number of true objects to be an unknown random variable. The uncertainty represented by the RFS may or may not further decompose into a one-to-one correspondence of internal components to actual objects. This latter situation describes the early Soviet flow-based trackers \cite{ClarkNarykovStreit2023} and their RFS antecedent, the Probability Hypothesis Density (PHD) filter \cite{MahlerBook1, phd-filter}. PHD-like filters consider RFSs on the state space only and estimate the spatial intensity of target occurrence. As a result, additional techniques are  required to extract per-target state estimates from the intensity function. By contrast, labeled RFS trackers model both the state space and a discrete label space, associating distinct labels with each hypothesized object. Labeled Multi-Bernoulli (LMB) filters \cite{LRFS1} directly associate labels with state-space components, while Generalized LMB (GLMB) filters \cite{vo2014labeled} represent a set of multi-object label hypotheses. GLMBs can thus be thought of as probabilistic MHTs with a more principled method of hypothesis probability generation. 

As noted in \cite{WilliamsMaMB2015}, these RFS frameworks require full enumeration of all one-to-one mappings between detections and hypothesized objects. Speedups have been obtained through $K$-best methods as noted above, clever merging of non-persistent hypotheses \cite{HoangVoVoFastGLMB}, and partitioned Gibbs sampling of hypothesis sets \cite{Gibbs_Efficient_2017, BeardVoVo2020}. Additionally, GPU-accelerated versions of some tracking algorithm components such as $K$-best matching \cite{HyLAC, BlockCUDAHungarian} are available, though these would still require careful tuning. 

We also note that in the multi-sensor setting, the standard GLMB model \cite{LRFS1, Multi_sensor_2019}  assumes that (1) each object generates at most one detection per sensor, and (2) each measurement originates from at most one object. While extended Target Tracking (ETT) generalizations of GLMB models \cite{ETT_2016} are capable of accommodating multiple detections per object, they typically do so by introducing models of object geometry and considering the joint assignment of measurement subsets. This again requires the use of $K$-best methods \cite{ETT_2016} or Gibbs sampling approximations \cite{Gibbs_Efficient_2017} that are not amenable to acceleration via parallel computing. Our approach addresses the both the GLMB filter's problem with multiple detections per object/sensor and its lack of of parallel scalability.

\section{Approach}
\label{approach}
RFS methods for MOT represent the multi-object state at each time $k$ as a random finite set describing the kinematic states of a collection of $N_k$ objects \(X_k = \{x_{1,k},...,x_{N_k,k}\}\). The kinematic states \(x_{i,k}\) are themselves random vectors. We note that the dimensionality of the kinematic state depends on the specification of an object dynamics model. We focus on vehicle tracking using a non-linear, five-dimensional constant turn angle and velocity dynamics model \cite{polack2017kinematic}.

To maintain track identity over time, the GLMB filter \cite{LRFS1} uses labeled RFSs to model the multi-object state. Each object state is augmented with a unique and persistent label \(l\in \mathbb{L}\), where \(\mathbb{L}\) is a discrete label space. A labeled state is a pair \(\mathbf{x}=(x,l)\), where \(x\) is the kinematic state and \(l\) is the label. We let \(\mathbf{X}_{k} = \{\mathbf{x}_{1,k},...,\mathbf{x}_{N_k,k}\}\) represent a labeled multi-object state at time $k$. Importantly, the labels in the labeled multi-object state must all be unique for the state to be well defined, but this property is trivial to enforce by construction \cite{LRFS1}. 

In RFS methods, the set of detections or measurements at time $k$ is also modeled as an RFS and is denoted by \(Z_k = \{z_{1,k},...,z_{M_k,k}\}\) where  $z_{i,k}$ denotes the $i^{th}$ measurement at time $k$ and the number of measurements at time $k$ is \(M_k\). 

The problem of interest in multi-object tracking is to infer the posterior distribution over the multi-object state $\mathbf{X}_k$ at time $k$ given the multi-object measurement history \(Z_{1:k}\) up to time $k$: \(\pi_{k}(\mathbf{X}_k|Z_{1:k})\). To accomplish this, the posterior from time 
\(k-1\), $\pi_{k-1}(\mathbf{X}_{k-1}|Z_{1:k-1})$,
is predicted forward to time \(k\) via the Chapman-Kolmogorov equation yielding the predicted density $\pi_{k|k-1}(\mathbf{X}_k|Z_{1:k-1})$ \cite{LRFS1}.
%
Given the new measurement set \(Z_k\) at time $k$, the predicted density is updated using Bayes' rule. Let  \(g_k(.|.)\) be the multi-object likelihood function, which models the measurement process, including detections, clutter, and sensor noise. The full posterior update is given by Equation \ref{eq:posterior}. However, computing this integral exactly is not feasible in general, particularly for non-linear dynamics.
\begin{align}
    \label{eq:posterior}
    \pi_k(\mathbf{X}_k|Z_{1:k}) = 
            \frac{g_k(Z_k|\mathbf{X}_k)\pi_{k|k-1}(\mathbf{X}_k)}
            {\int g_k(Z_k|\mathbf{X})\pi_{k|k-1}(\mathbf{X})\delta \mathbf{X}}
\end{align}

At each iteration, our filter begins with an approximation to the full multi-object posterior \(\pi_{k-1}(\mathbf{X}_{k-1}|Z_{1:k-1})\) from the previous time step, which is a GLMB density as formulated in \cite{LRFS1}. 
To accommodate non-linear dynamics, we use a weighted particle representation of the probability distribution over each labeled state \cite {bpfgordon1993novel} \cite{bpfarulampalam2002}. Specifically, state \(\mathbf{x}_{i,k}\) is approximated by a set of \(P\) weighted particles, \(\{(w_{i,k}^{p}, \mathbf{x}_{i,k}^{p})\}\), where \(\mathbf{x}_{i,k}^{p}\) and  \(w_{i,k}^{p}\) are the labeled kinematic state and corresponding weight of the \(p^{th}\) particle .

Our prediction step models target dynamics using a standard birth and survival framework \cite{LRFS1}, \cite{ETT_2016}. New potential targets are introduced via a Labeled Multi-Bernoulli (LMB) birth process, where a set of particles is sampled from the birth probability distribution for each new track. Simultaneously, particles representing targets that have survived from step \(k-1\) are propagated forward using the constant turn rate and velocity kinematics model \cite{polack2017kinematic}. This process can be performed as a batch operation over all particles and surviving tracks. The complete predicted density is formed by the superposition of these surviving predicted and newborn tracks distributions. Let the total number of tracks at this point be \(T_k\) and their multi-object state be denoted as \(\mathbf{X}_k^+\).

We next turn to the problem of computing a set of track-to-measurement association hypotheses. We begin by constructing an \(M_k \times (T_k+1)\) compatibility matrix, \(C_k\) from the set of all \(M_k\) measurements and \(T_k\) predicted tracks:
\begin{align}
C_k[i,j] =
\begin{cases}
   \begin{aligned}[b]
       &  g_{ijk} \cdot L(z_{i,k}|\mathbf{x}_{j,k})
   \end{aligned}
   & \text{for } j = 1, \dots, T_k \\
   \kappa & \text{for } j = 0 \text{ (clutter)}
\end{cases}
\end{align}
The constant $\kappa$ represents the clutter probability while $P_D$ represents the detection probability. $L(z_{i,k}|\mathbf{x}_{j,k})$ represents the likelihood of measurement $i$ under the probability distribution defined by the state $\mathbf{x}_{j,k}$ of predicted object $j$. The terms $g_{ijk}$ represent binary indicators for the output of a standard gating filter $g_{ijk} = L(z_{i,k}|\mathbf{x}_{j,k})\geq \gamma$ that helps to reduce the complexity of the association step by completely zeroing out unlikely associations \cite{barshalom1988tracking}. 
For $j>0$, the values $C_k[i,j]$ represents the probability that measurement $i$ is not clutter, that hypothesized object $j$ was detected, and that measurement $i$ was generated by hypothesized object $j$. For $j=0$, $C_k[i,j]$ represents the probability that measurement $i$ is clutter. 
Importantly, we note that the entire construction of the joint compatibility  matrix can be performed in parallel.


Once the compatibility matrix is constructed, we use it to update the measurement-to-object association hypotheses. We let $\theta_k$ denote the association map between track labels at time $k$  and measurement indices at time $k$ similar to the formulation in \cite{LRFS1}. We then let $\xi_k = (\theta_1, ..., \theta_{k})$ represent the history of associations up to time $k$. Let the pair \((\mathbf{X}_k,\xi_k)\) denote the hypothesis that the set of tracks $\mathbf{X}_k$ has the history of association maps specified by \(\xi_k\). We let \(w^{(\mathbf{X}_k,\xi_k)}\) represent the probability of hypothesis \((\mathbf{X}_k,\xi_k)\). 

For each of the \(H_{old}\) hypotheses from the previous timestep $k-1$, a new set of \(H_{up}\) hypotheses is generated for the current time step. This process begins by determining the survival or death of each track $\mathbf{x}_{j,k}$ within a given hypothesis. We compute a posterior probability of track death for each track $d_{prob}(\mathbf{x}_{j,k})$, balancing its prior survival probability against the likelihood of it being detected, similar to the approach in \cite{Multi_sensor_2019}. 
%
 
For each existing hypothesis, new successor hypotheses are generated through a two stage sampling process. First, we perform a Bernoulli sampling step using $d_{prob}(\mathbf{x}_{j,k})$ to determine which tracks survive. The columns in the compatibility matrix \(C_k\) corresponding to tracks that were terminated are zeroed out yielding a modified matrix \(C'_k\). Then, for each measurement $i$, an association is sampled from a categorical distribution defined by normalizing the corresponding row in the modified matrix \(C'_k\), assigning it either to a surviving track $j$ or to clutter. We run this two stage sampling process multiple times per old hypothesis yielding a large set of new assignment hypotheses from which only the unique hypotheses are retained. 

Importantly, while the absence of constraints on the number of detections generated per track (or per track-sensor pair) in our approach means that the space of all possible assignments is exponentially large (its precise size is $(T_k+1)^{M_k}$), the measurement-to-track associations are only coupled through cardinality priors on the number of measurements per object. We can ignore the cardinality priors, sample the associations from a completely independent proposal distribution , and then apply importance weights. The complete hypothesis generation procedure and weighting step can then be parallelized across the batch of prior hypotheses. 

This is in contrast to the standard approach in \cite{LRFS1} and \cite{Multi_sensor_2019} where the assumption that each track generates at most one measurement results in a smaller space of associations than our approach. However, the space of associations in the standard approach is still combinatorially large and requires the application of sequential sampling methods such as the Gibbs sampler due to hard limits on the number of measurements per track.
As noted previously, such sequential sampling algorithms are not amenable to parallel computation. 

Finally, we perform a batched particle update for all unique (track, measurement) association pairs generated across the new set of hypotheses. The particle weights are updated according to the measurement likelihood, and a resampling step is performed if necessary to prevent particle degeneracy. The weight of each new hypothesis is calculated by multiplying its parent's weight with the joint likelihood of the sampled survival/death events and data associations. Finally, to prevent exponential growth, the hypothesis set is pruned. First, hypotheses with weights below a predefined threshold $\tau$ are discarded. Finally, the \(H_{max}\) most likely hypotheses are retained and the remainder are discarded.

As noted throughout this description, all steps in the algorithm implementing the update step for the proposed tracker can be computed in parallel over hypotheses, measurements and tracks. This makes the algorithm highly amenable to acceleration using GPU computing. To flexibly take advantage of GPU computing resources while also allowing the implementation to run on a standard CPU, we implement the tracker in Python using the PyTorch library \cite{paszke2019pytorch}. In the next section, we present an initial evaluation of the approach.

\begin{figure}[t]
    \centering
    \includegraphics[width=\linewidth]{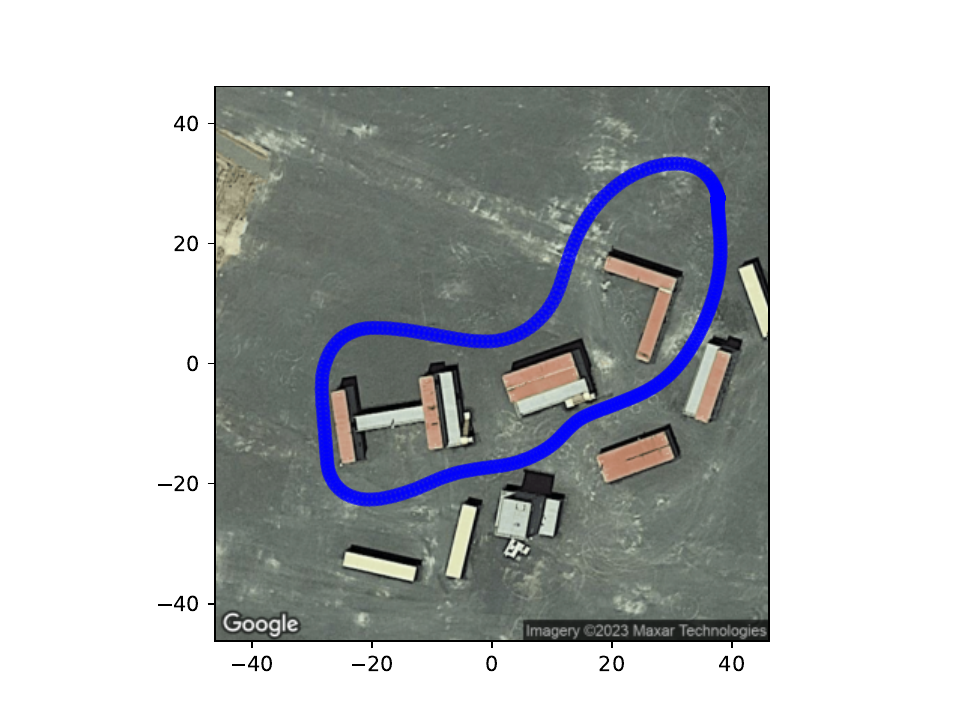}
    \caption{Visualization of testbed ground truth track.}
    \label{ground_truth_track}
    \vspace{-1em}
\end{figure}

\begin{figure*}[t!]
    \centering
    \begin{subfigure}[t]{0.24\textwidth}
        \includegraphics[width=\textwidth]{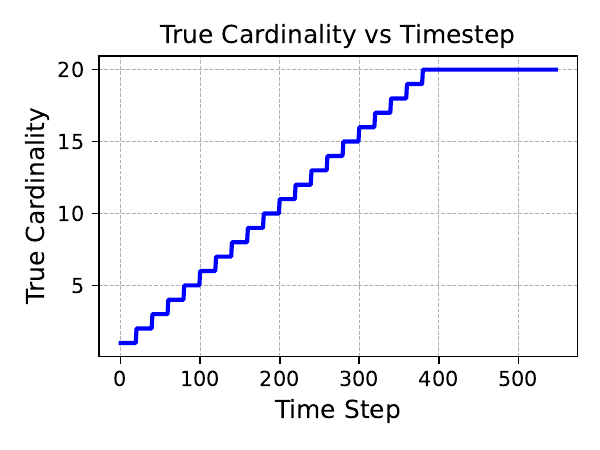}
        \subcaption{True Cardinality}
    \end{subfigure}
    \hfill
    \begin{subfigure}[t]{0.24\textwidth}
        \includegraphics[width=\textwidth]{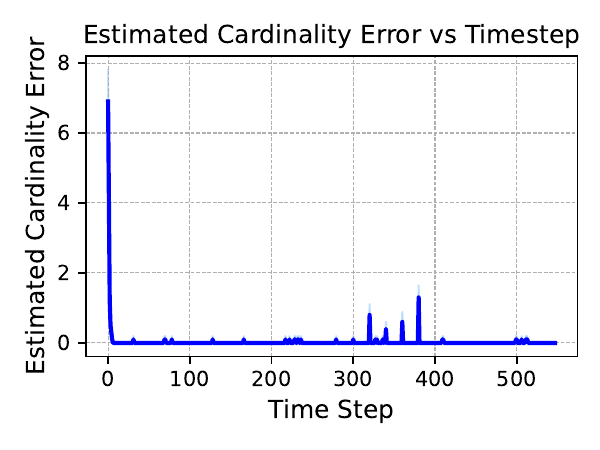}
        \subcaption{Estimated Cardinality Error}
    \end{subfigure}
    \hfill
    \begin{subfigure}[t]{0.24\textwidth}
        \includegraphics[width=\textwidth]{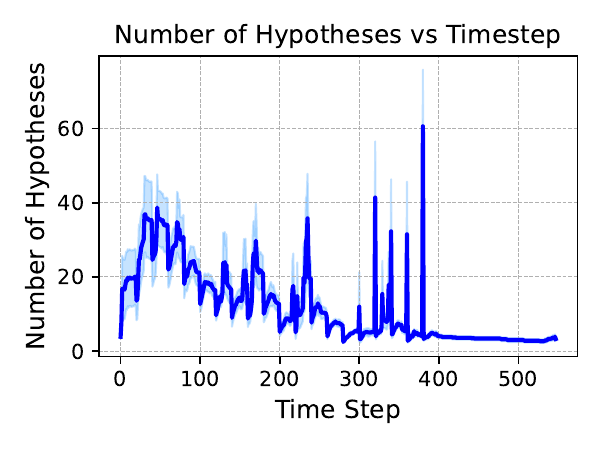}
        \subcaption{Number of Hypotheses}
    \end{subfigure}
    \hfill
    \begin{subfigure}[t]{0.24\textwidth}
        \includegraphics[width=\textwidth]{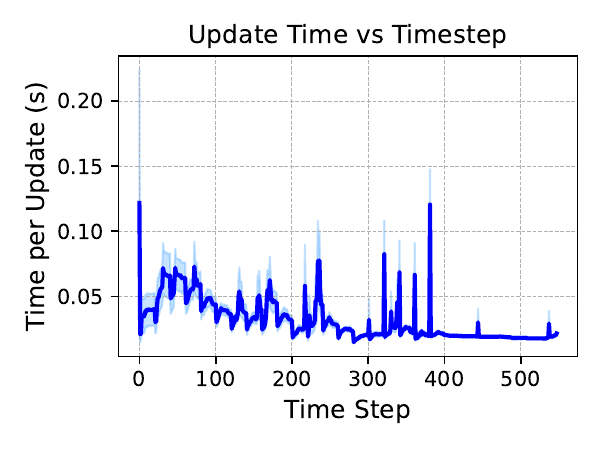}
        \subcaption{Update Time}
    \end{subfigure}
    \caption{These plots show metrics for the case of 20 objects, a maximum of $100$ hypotheses, and the L40S GPU. (a) shows the true cardinality (e.g. the true number of objects in the environment) at each time point. (b) shows the cardinality error of the tracker. (c) shows the number of hypotheses at each time point. (d) shows the update run time at each time point. For (a)-(c) the results are the mean of 10 runs and the shaded region is the 
    standard error of the mean.} 
    \label{fig:vis}
\end{figure*}

\begin{figure*}[t!]
    \centering
    \begin{subfigure}[b]{0.24\textwidth}
        \includegraphics[width=\textwidth]{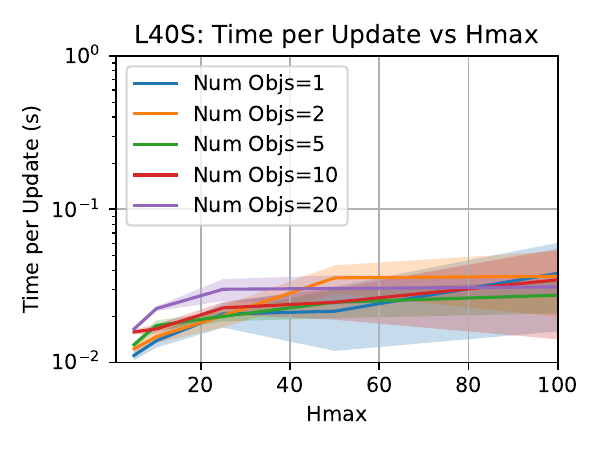}
        \subcaption{Nvidia L40S\\ (Server GPU)}
    \end{subfigure}
    \begin{subfigure}[b]{0.24\textwidth}
        \includegraphics[width=\textwidth]{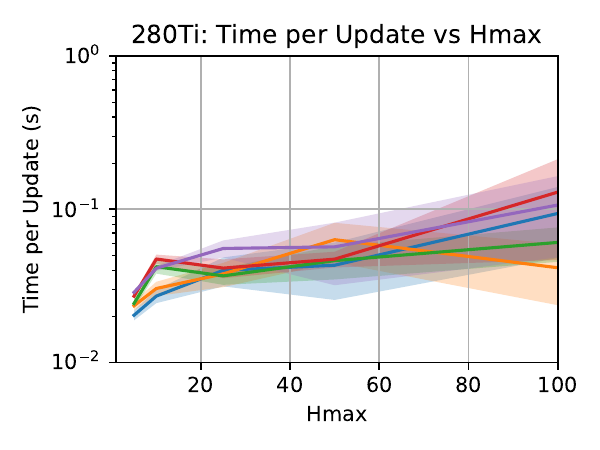}
        \subcaption{Nvidia 2080Ti\\ (Server GPU)}
    \end{subfigure}
    \begin{subfigure}[b]{0.24\textwidth}
        \includegraphics[width=\textwidth]{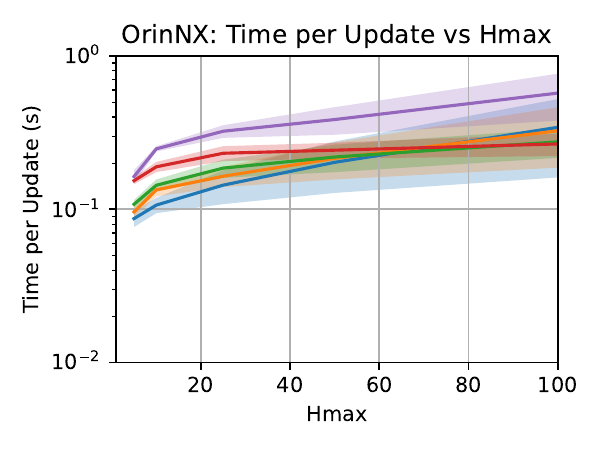}
        \subcaption{Nvidia Orin NX\\ (Edge GPU)}
    \end{subfigure}
    \begin{subfigure}[b]{0.24\textwidth}
        \includegraphics[width=\textwidth]{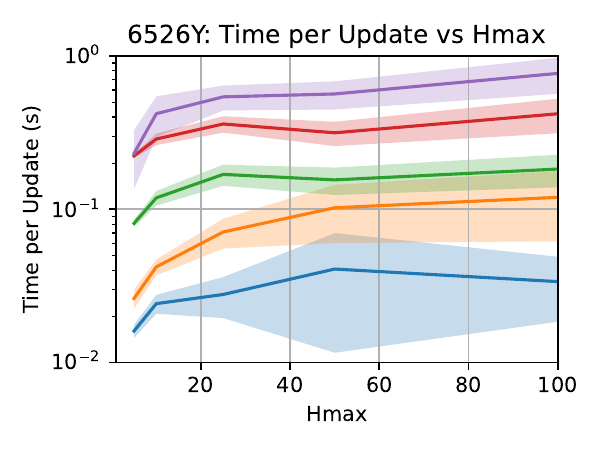}
        \subcaption{Intel 6526Y\\ (Server CPU)}
    \end{subfigure}
    \caption{These plots show the per-update mean run time of GPU-GLMB on several hardware platforms as described in Table 1. Each line shows the per-update mean run time for a different number of ground truth objects. The shaded regions correspond to one standard error of the mean. } 
    \label{fig:scaling}
\end{figure*}

\begin{figure}[t!]
    \centering
    \begin{subfigure}[b]{0.48\linewidth}
        \includegraphics[width=\textwidth]{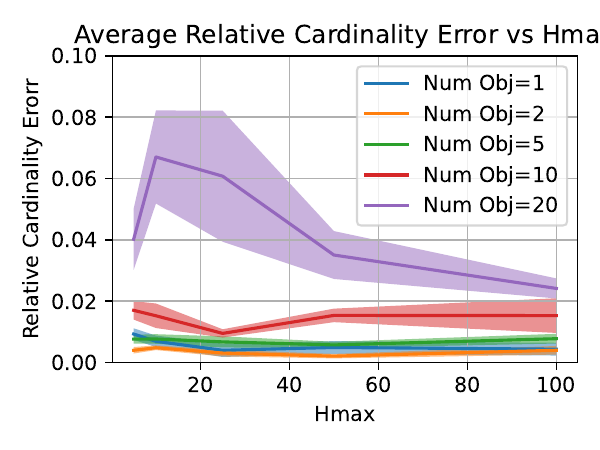}
        \subcaption{Cardinality Error}
    \end{subfigure}
    \begin{subfigure}[b]{0.48\linewidth}
        \includegraphics[width=\textwidth]{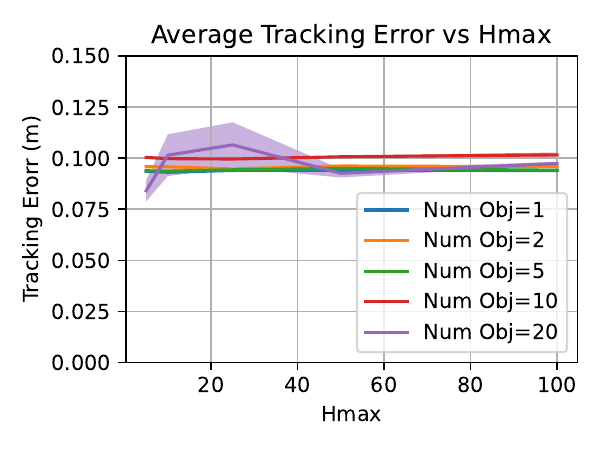}
        \subcaption{Tracking Error}
    \end{subfigure}
    \caption{These plots show the average cardinality and tacking
    errors as a function of the number of objects and maximum number of hypotheses
    computed using the L40S GPU.} 
    \label{fig:performance}
        \vspace{-1.5em}
\end{figure}

\section{Experiments and Results}
\label{experiments}

\subsection{Experimental Protocols}

The computational scalability of multi-hypothesis trackers depends on a number of parameters of the tracker as well as properties of the tracking problem. Key parameters of the GLMB tracker itself include the maximum number of hypotheses $H_{max}$, the number of new hypotheses sampled during updating $H_{up}$, and the hypothesis pruning probability threshold $\tau$. For the tracking problem, the key scaling parameters are the true number of objects $N$, the expected number of detections per object $D$, and the clutter spatial density $\lambda$. 

To aid in assessing the scalability of GPU-GLMB, we create a benchmark of tracking problems where we can control these key parameters. We begin with a ground truth single-object track collected using a high precision GPS system. This track was collected using the IoBT-MAX deployment at the DEVCOM Army Research Laboratory's Robotics Research Collaboration Campus (R2C2) \cite{marlin2023iobt}. We show a visualization of the track in Figure \ref{ground_truth_track}. Let $\mbf{x}^*=[x^*_0,...,x^*_K]$ be the ground truth position observations for this object. We first make $N$ copies of the ground truth track where each copy $n$ is offset in time relative to the base object. Specifically, we let $\mbf{x}^*_n=[x^*_{n0},...,x^*_{nK}]$ where $x^*_{n,k} = x^*_{k'}$ with $k'=\mbox{mod}(n\delta, K)$. This has the effect of creating a simulated convoy of objects that all follow the same trajectory but where each copy is offset in time from the previous copy by $\delta$ time steps. 

Next, to create the detections at time $k$ for each copy $n$, we independently sample $D$ mean vectors $\mu_{ndk}\sim\mathcal{N}(x^*_{n,k},\sigma^2I)$. For simplicity, we use the same detection covariance matrix $R_{ndk}=\sigma^2I$ for all copies, time points and detections. Here $\sigma^2$ controls both the noise in the detection means and the detection uncertainty. While we leave the detections in sorted order, we note that the tracker is invariant to the order of the detections within a single update. Each timestep corresponds to 0.1 seconds. The length of the testbed scenario is 54.8s.

To assess the scalability of the tracker, we focus on the number of objects $N$ and the maximum number of hypotheses $H_{max}$. We perform experiments using $N\in\{1,2,5,10,20\}$ and  $H_{max}\in\{5,10,25, 50,100\}$. We fix $D=5$, $\lambda=0$, $\sigma=0.25$ and $\delta=2$, $H_{up}=100$, and $\tau=10^{-5}$. We re-run each configuration of the tracker on each tracking problem configuration 10 times. We report results in terms of the mean and standard error over the 10 experiment repetitions. 

Lastly, we provide results for several compute platforms, as described in Table I. We control the device on which GPU-GLMB computations are performed by setting PyTorch's ``device" attribute to ``cpu" for CPU-only computation or ``cuda" for GPU-accelerated computation. In total across all platforms and experiment configurations, 1000 runs of GPU-GLMB were performed.


\begin{table}[t!]
    \centering
    \caption{Hardware configurations for scaling experiments.}
    \resizebox{\columnwidth}{!}{
    \def\arraystretch{1.25}
    \begin{tabular}{|c|c|r|r|}\hline
         Processor       & Type       & FP32 TFLOPS & VRAM (GB)  \\\hline\hline
         Intel 6526Y     & Server CPU & 0.2      & -  \\\hline
         Nvidia Orin NX  & Edge GPU   & 1.9      & 16.0  \\\hline
         Nvidia 2080Ti   & Desktop GPU & 13.4    & 11.0  \\\hline
         Nvidia L40S     & Server GPU & 91.6     & 40.0  \\\hline
    \end{tabular}%
    } 
    \label{tab:placeholder}
\end{table}

\subsection{Results}

In Figure \ref{fig:vis} we show example output from GPU-GLMB for the most computationally intensive case considered ($N=20$ objects and $H_{max}=100$). We visualize the true cardinality of the objects in the environment, the absolute cardinality error, the number of hypotheses, and the GPU-GLMB tracker update run time per time step. All plots show the mean over 10 runs as a solid line and the standard error of the mean as a filled region. We can see that except for the initial time points, the cardinality error is close to zero. We can also see that the number of active hypotheses stays under the maximum threshold of 100. Finally, we can see that in this configuration, the GPU-GLMB tracker has an average update time that is under the threshold of 0.1 seconds per update needed to maintain real-time performance almost always.

We provide an overview of tracker accuracy in Figure \ref{fig:performance}. These results were computed using the L40S GPU (since all compute platforms are running the same implementation, the tracking performance results are essentially identical on all platforms). In Figure \ref{fig:performance}(a), we show the average relative absolute cardinality error. To compute this metric, we take the absolute difference between the true number of objects and estimated number of objects and normalize by the true number of objects at each time point in each run of the tracker. We then average those values across time points for each run. Finally, we compute the mean and standard error of the mean across runs for each number of objects and each value of $H_{max}$. We can see that the relative cardinality error is less than 2\% for up to 10 objects. It increases to $10\%$ when using 20 objects and an insufficiently large value of $H_{max}$, but returns to a low value when using 100 hypotheses.  

Next, Figure \ref{fig:performance}(b) shows the average tracking error per object in meters. To compute this metric, we apply the Hungarian algorithm at each timestep of each run to compute the minimum distance matching between the set of predicted object locations output by the tracker to the set of ground truth object locations. We then compute the average distance between true and predicted object locations for the optimal matching. We repeat this process for all time points and then compute the time average of the tracking error per run. We visualize the mean and standard error of the mean across runs for this metric for each number of objects and each value of $H_{max}$. We can see that the average tracking error for matched objects is less than 0.15m for all configurations of the tracker relative to a tracking area of $80\times 80$ meters (see Figure \ref{fig:vis}).

We next turn to a discussion of the computational scalability of the GPU-GLMB across different configurations of the tracking problem, the tracker, and the compute hardware used. The metric that we assess is the average time per GPU-GLMB tracker update in seconds. We compute the average update time per run and report the mean and standard error of the mean over runs for each number of objects and each value of $H_{max}$. The results are shown in Figure \ref{fig:scaling}. Note that the graphs use a log y-axis to more clearly show trends in the results given the high spread in run time values. The first major trend that we can see is that the server GPUs significantly outperform the server CPU in scenarios with $5$ or more objects. Indeed, the average update time for the server CPU exceeds the real time limit of 0.1 seconds for all values of $H_{max}$ when using $5$ or more objects. Next, we can see that between the the server and desktop GPUs, the more powerful L40s GPU exhibits very little scaling in update run time with respect to the value of $H_{max}$. By contrast, we can see that the less powerful 2080Ti begins to see slower update times beyond $H_{max}=50$. Finally, the edge GPU is only able to achieve real time update performance for a smaller number of objects and $H_{max}=5$ where Figure \ref{fig:performance}(a) shows increased cardinality errors. Interestingly, while the edge GPU has slower update times than the server CPU for smaller numbers of objects, its run time becomes competitive with the server CPU for larger numbers of objects.
\section{Conclusions and Future Work}
\label{conclusion}
In this paper, we have presented a modification of the standard GLMB filter that can accommodate multiple detections per target without introducing explicit models of object geometry. We have shown that this change to the standard GLMB model can be leveraged to to develop a fully vectorized implementation of a multi-sensor, multi-target tracker in Python using the PyTorch library. 

Our scalability results indicate that GPU-GLMB implementation can successfully leverage the parallel computing capability of  server GPUs such as the Nvidia L40S to maintain faster than real time performance across all experiment settings studied. These results indicate the potential for GPU-GLMB to be incorporated into a heterogeneous, bandwidth limited compute infrastructure where lower power edge nodes are used for sensing, computing and transmitting detections to a server-class GPU fusion center running GPU-GLMB. In future work, We plan to continue to optimize the GPU-GLMB implementation and to investigate scaling to larger settings using more complex multi-object and multi-sensor data.

\section*{Acknowledgment}
This research was sponsored by the Army Research Laboratory and was accomplished under Cooperative Agreement Number W911NF-17-2-0196.  The views and conclusions contained in this document are those of the authors and should not be interpreted as representing the official policies, either expressed or implied, of the Army Research Laboratory or the U.S. Government. The U.S. Government is authorized to reproduce and distribute reprints for Government purposes notwithstanding any copyright notation herein.

\bibliographystyle{plain} 
\bibliography{bib}

\end{document}